\documentclass[11pt]{article}

\usepackage[preprint]{acl}

\usepackage{times}
\usepackage{latexsym}

\usepackage[T1]{fontenc}

\usepackage[utf8]{inputenc}

\usepackage{microtype}

\usepackage{inconsolata}

\usepackage{graphicx}
\usepackage{amssymb}

\usepackage[utf8]{inputenc} 
\usepackage[T1]{fontenc}    
\usepackage{nicefrac}       
\usepackage{microtype}      

\usepackage{amsmath,amsfonts}
\usepackage{algorithmic}
\usepackage{algorithm}
\usepackage{textcomp}
\usepackage{stfloats}
\usepackage{url}
\usepackage{verbatim}
\usepackage{tabularx, xcolor, colortbl, booktabs, multirow, array} %
\newcolumntype{C}{>{\centering\arraybackslash}X} %
\usepackage[misc]{ifsym}
\usepackage[table]{xcolor}
\usepackage{makecell}
\usepackage{threeparttable}
\usepackage{pifont}
\usepackage{subcaption}  
\usepackage{hyperref}

\usepackage{amsthm}

\usepackage{bm}
\usepackage{bbm}

\usepackage{tcolorbox}
\tcbuselibrary{listings, breakable}

\usepackage{listings}
\newtcolorbox{promptbox}{
  breakable,
  colback=gray!3,
  colframe=gray!40,
  boxrule=0.4pt,
  arc=2pt,
  left=6pt,
  right=6pt,
  top=6pt,
  bottom=6pt
}

\definecolor{linkblue}{RGB}{0, 92, 175}

\hypersetup{
    colorlinks=true,
    linkcolor=linkblue,
    citecolor=linkblue,
    urlcolor=linkblue
}

\author{
Chaodong Tong\textsuperscript{1,2}
\quad
Qi Zhang\textsuperscript{3}
\quad
Zhuojun Jiang\textsuperscript{1}
\quad
Lei Jiang\textsuperscript{1}
\quad
Yanbing Liu\textsuperscript{1,2}
\\[0.5em]
\textsuperscript{1}Institute of Information Engineering, Chinese Academy of Sciences, Beijing, China
\\
\textsuperscript{2}School of Cyber Security, University of Chinese Academy of Sciences, Beijing, China
\\
\textsuperscript{3}China Industrial Control Systems Cyber Emergency Response Team, Beijing, China
\\[0.5em]
\textsuperscript{1}\texttt{\{tongchaodong, jiangzhuojun, jianglei, liuyanbing\}@iie.ac.cn}
\\
\textsuperscript{3}\texttt{bonniezhangqi@126.com}
}

\title{HaluNet: Learning Hallucination Risk from Internal Signals in LLM Question Answering}

\begin{document}

\maketitle

\begin{abstract}
Large language models (LLMs) achieve strong question answering (QA) performance
but can produce fluent answers unsupported by available evidence. Existing
hallucination detectors often rely on external verification, repeated sampling,
or test-time judge calls, which can be costly for real-time QA. We propose
\textbf{HaluNet}, a lightweight hallucination risk estimator that uses internal
signals from one model generation. HaluNet jointly models token likelihood,
predictive entropy, and hidden-state information, allowing probabilistic,
distributional, and semantic evidence to inform an answer-level risk score. It
is trained with LLM-as-a-Judge labels as scalable weak supervision and evaluated
with independent human and multi-judge assessments. Experiments on SQuAD,
TriviaQA, and Natural Questions show that HaluNet improves answer-level risk
ranking across in-domain and out-of-domain settings. On a 300-example human
evaluation, HaluNet achieves 0.874 AUROC and 0.869 AUPRC; its top 20\%
highest-risk answers contain 96.5\% errors, yielding a 2.06$\times$ lift over
the base error rate.
\end{abstract}

\section{Introduction}
Question answering (QA) is a central capability of modern large language models
(LLMs), supporting applications from search engines and conversational assistants
to autonomous LLM agents. Despite strong open-domain QA performance
~\cite{tan2023can}, LLMs can generate hallucinated
answers with factual errors or unsupported claims, undermining reliability
~\cite{farquhar2024detecting,ji2023survey}. Verification against external
knowledge sources or through multi-step checking can mitigate this risk, but it
often incurs substantial inference cost and depends on knowledge coverage
~\cite{cheung2023factllama,ge2025resolving,wang2023knowledge}. This motivates a
complementary question: can the reliability risk of a generated answer be
estimated directly from internal signals of a single generation, without
repeated sampling or test-time verification?

Hallucination in LLMs has been characterized as factual inconsistency,
unsupported generation, fabrication, or weak attribution to available evidence
~\cite{ji2023survey,li2023halueval,ravichander2025halogen}. In QA applications,
these phenomena converge on a practical reliability problem: a system may return
a plausible answer that lacks sufficient factual support. We therefore study
\emph{answer-level hallucination detection}, where a generated QA answer is
considered hallucinated when it is contradicted by, or cannot be verified
against, the available context, reference answer, or factual knowledge.

\begin{figure}[t]
\centering
\includegraphics[width=\columnwidth]{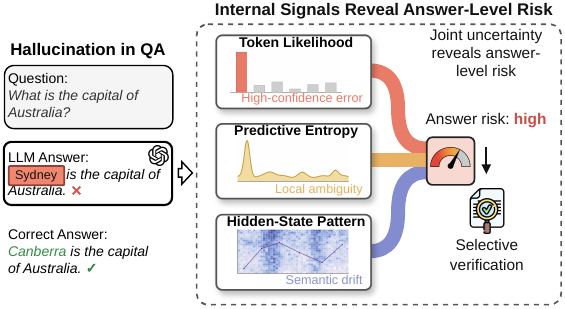}
\caption{
Individual internal signals provide incomplete evidence for QA hallucination risk,
whereas token likelihood, predictive entropy, and hidden-state patterns capture
complementary aspects of a single generation trace.
}
\label{fig:overview}
\end{figure}

Existing hallucination detectors often rely on external verification,
sampling-based uncertainty estimation, or internal generation signals.
Verification-based methods compare answers with retrieved documents, knowledge
bases, or model-based judges, while sampling-based methods measure uncertainty
or semantic consistency across multiple generations
~\cite{kossen2024semantic,manakul2023selfcheckgpt,farquhar2024detecting}. These
approaches can provide useful reliability evidence but require additional
retrieval, verification, judge calls, or repeated decoding. Internal signals
such as token likelihoods, predictive entropy, and hidden states are available
from the model's own generation process and therefore offer a promising basis for
low-overhead risk estimation
~\cite{grewal2024improving,jiang2021can,liu2022token,zhang2023enhancing,
aryal2025howard,zhang2025detecting,guo2025reward}. However, existing methods
often rely on a limited subset of signals or treat them as separate summary
features, leaving open how to jointly exploit their evidence for answer-level
risk estimation.

To address this gap, we propose \textbf{HaluNet}, a lightweight framework that
estimates answer-level risk from one model generation
(Fig.~\ref{fig:overview}). HaluNet is based on the observation that different
internal signals reflect different aspects of answer reliability: token
likelihood captures confidence in the emitted path, predictive entropy measures
uncertainty over alternative continuations, and hidden states provide
generation-conditioned semantic evidence. Rather than treating any signal as a
standalone indicator, HaluNet learns a joint risk representation that integrates
probabilistic, distributional, and semantic evidence. The model is trained with
LLM-as-a-Judge labels as scalable, model-generated weak supervision, while
inference uses only the internal signals of the given generation.

Our contributions are summarized as follows:
\begin{itemize}
\item We present \textbf{HaluNet}, a hallucination risk estimator that jointly
models probabilistic, distributional, and hidden-state evidence to assess
answer-level reliability.
\item We train HaluNet with judge-generated weak supervision and validate the
learned risk scores through a 300-example human evaluation and an independent
multi-judge assessment.
\item We show that HaluNet achieves strong answer-level risk ranking across
in-domain, out-of-domain, and human-evaluated settings, while avoiding
test-time judge calls, repeated sampling, and post-hoc verification.
\end{itemize}

\section{Related Work}
\paragraph{Hallucination scope in QA.}
Hallucination is used both as a broad reliability term and as a
task-specific notion of unsupported generation. QA benchmarks typically label an
answer as hallucinated when it contradicts the source context or cannot be
verified by factual knowledge~\cite{li2023halueval,ravichander2025halogen}.
Although some studies distinguish causal sources of unsupported outputs
~\cite{ji2023survey}, our work follows the evaluation-oriented
QA setting: a generated answer is treated as hallucinated when it is
contradicted by, or lacks support from, the available context, reference answer,
or factual knowledge.

\paragraph{Verification and sampling-based detection.}
A major line of work checks generated content against retrieved evidence,
external knowledge bases, or model-based judges
~\cite{wang2023knowledge,cheung2023factllama,ge2025resolving}. Another line
estimates uncertainty or semantic consistency from multiple generations
~\cite{manakul2023selfcheckgpt,farquhar2024detecting,kossen2024semantic}.
These methods provide useful evidence for factual support or semantic
uncertainty, but they require additional retrieval, judging, or repeated
decoding. HaluNet targets a complementary setting: assigning risk to a generated
QA answer using signals already produced by the generation itself.

\paragraph{Internal-signal and learned detectors.}
Internal-signal methods use information exposed during generation, including
token likelihoods, logits, entropy, hidden states, attention patterns, and
semantic embeddings~\cite{liu2022token,zhang2023enhancing,grewal2024improving,
guo2025reward,zhang2025detecting}. Recent learned detectors further map
model-derived reliability features to risk or calibration scores
~\cite{liu2024uncertainty,yaldiz-etal-2025-design}. These studies show the
value of internal evidence, but many methods emphasize one dominant signal
family or aggregate internal behavior into global features. In contrast,
\textbf{HaluNet} jointly uses token likelihood, predictive entropy, and
hidden-state information from one QA generation to estimate answer-level risk
without test-time sampling, judge calls, or external verification.

\section{Methodology}
\begin{figure*}[t]
\centering
\includegraphics[width=\textwidth]{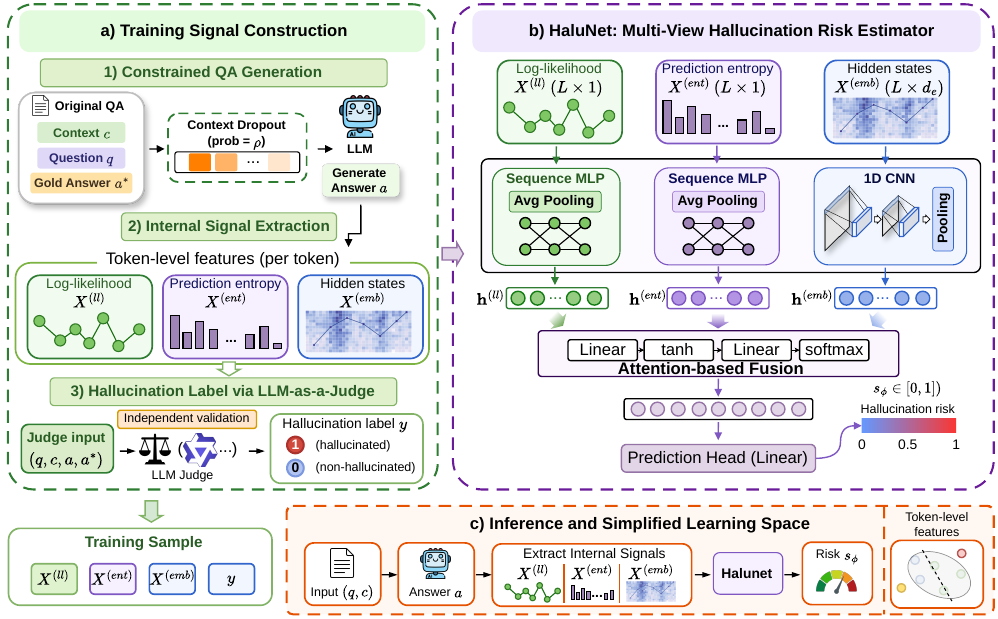}
\caption{
Overview of HaluNet.
The training pipeline constructs weakly labeled QA examples by generating
answers, extracting token-level internal signals, and assigning hallucination
labels with an LLM-as-a-Judge protocol.
HaluNet models log-likelihood, predictive entropy, and hidden states as three
views of the same generation and outputs an answer-level hallucination risk
score.
At inference time, the trained detector uses only internal signals from the
given generation.
}
\label{fig:halunet}
\end{figure*}

\subsection{Task and Risk-Score Formulation}
\label{sec:training_data_construction}

We study answer-level hallucination detection for question answering. Given a
question $q$, an optional context $c$, and an answer $a$ generated by a backbone
language model $\mathcal{M}_{\theta}$, the goal is to assign a scalar risk score
indicating whether $a$ is contradicted by, or unsupported under, the reference
answer or the available evidence. Let $a=(x_1,\ldots,x_T)$ denote the generated
token sequence, and let
\begin{equation}
\mathcal{X}(q,c,a)=
\{X^{(\mathrm{ll})},X^{(\mathrm{ent})},X^{(\mathrm{emb})}\}
\end{equation}
denote the internal signals extracted from the given generation. HaluNet
predicts
\begin{equation}
s_{\phi}(q,c,a)
=
P_{\phi}\!\left(y=1\mid \mathcal{X}(q,c,a)\right),
\end{equation}
where $y=1$ denotes an unsupported or hallucinated answer. A larger $s_{\phi}$ indicates higher answer-level hallucination risk. We treat
$s_{\phi}$ as a learned answer-level unreliability score; it is not intended to
serve as a calibrated uncertainty estimate in the sense of uncertainty
quantification or language-model calibration~\cite{hullermeier2021aleatoric,
jiang2021can,huang-etal-2024-uncertainty}.

Training labels are produced by an LLM-as-a-Judge protocol
\cite{li2023halueval,gu2024survey,farquhar2024detecting}.
For each tuple $(q,c,a)$, the judge receives the question, the generated answer,
the reference answer, and the context when available, and returns a binary
support decision. These labels are used as scalable weak supervision, while the
human and cross-judge evaluations in Section~\ref{sec:rq5_label_reliability}
assess their reliability. At inference time, HaluNet uses no judge calls or
additional sampled answers; it estimates risk only from the internal signals of
the given generation.

\subsection{Single-Pass Internal Signals}
\label{sec:feature_extraction}

HaluNet is built on the hypothesis that answer-level hallucination risk is often
not captured by any single internal signal. A generated answer may remain fluent
and high-confidence while lacking factual support, or uncertainty may concentrate
on only a few generated tokens. We therefore use three signals as different
views of the same generation as shown in Fig.~\ref{fig:halunet}. Log likelihood measures confidence along the
emitted token path. Predictive entropy is distributional, measuring how
probability mass is spread over alternative continuations. Hidden states are
representational, describing the generation-conditioned trajectory in the
backbone's semantic space. These views are related but not interchangeable: for
example, a model may assign high likelihood to an unsupported answer, or show
localized uncertainty without substantially changing the overall fluency of the
response. Appendix~\ref{sec:signal_correlation} provides an empirical analysis
of the correlations between internal signals and hallucination labels.

For each generated token $x_t$, we compute its log likelihood
\begin{equation}
\ell_t=\log p_{\mathcal{M}_{\theta}}(x_t\mid x_{<t},q,c),
\end{equation}
which captures the model's token-local confidence in what it actually
generated. We also compute the predictive entropy over the next-token
distribution,
\begin{equation}
\label{eq:token_entropy}
\begin{aligned}
H_t = -\sum_{v\in V}
&p_{\mathcal{M}_{\theta}}(v\mid x_{<t},q,c) \\
&\cdot \log p_{\mathcal{M}_{\theta}}(v\mid x_{<t},q,c).
\end{aligned}
\end{equation}
which captures whether the model assigns probability mass to a diffuse set of
plausible continuations. Finally, we record a hidden state
\begin{equation}
\mathbf{e}_t=\operatorname{Hidden}^{(l)}_{\mathcal{M}_{\theta}}
(x_t\mid x_{<t},q,c)\in\mathbb{R}^{d_e},
\end{equation}
which represents the generation-conditioned semantic trajectory of the answer.
In the main experiments, we use a fixed pre-specified intermediate layer for
feature extraction. The layer-band analysis in Appendix~\ref{sec:layer_sensitivity}
shows that intermediate representations are generally competitive, while fixing
one layer keeps feature extraction cost manageable and avoids selecting layers
on the human-evaluation or test sets.

Each answer is represented as
\begin{equation}
\mathcal{X}(q,c,a)=
\{X^{(\mathrm{ll})},X^{(\mathrm{ent})},X^{(\mathrm{emb})}\},
\end{equation}
where
$X^{(\mathrm{ll})},X^{(\mathrm{ent})}\in\mathbb{R}^{L\times 1}$
and $X^{(\mathrm{emb})}\in\mathbb{R}^{L\times d_e}$. Sequences are padded or
truncated to $L=50$ generated tokens. Appendix~\ref{sec:length_sensitivity} examines the effect of more aggressive truncation.

\subsection{Branch Encoding and Fusion}

HaluNet encodes each signal with a feature-specific branch. Let
$f\in\{\mathrm{ll},\mathrm{ent},\mathrm{emb}\}$ and
$X^{(f)}\in\mathbb{R}^{L\times d_f}$. We set the branch hidden size to
$d_h=64$ in all main experiments. The main architecture uses a mixed encoder.
For scalar probabilistic signals, log likelihood and entropy are first averaged
over the generated sequence and then projected with a small MLP:
\begin{align}
\bar{x}^{(f)}
&=\frac{1}{L}\sum_{t=1}^{L} X^{(f)}_t,
\quad f\in\{\mathrm{ll},\mathrm{ent}\},\\
h^{(f)}
&=\operatorname{MLP}_{\mathrm{avg}}^{(f)}(\bar{x}^{(f)}).
\end{align}
This branch captures the answer-level tendency of scalar confidence and
uncertainty signals without introducing extra sequence modeling capacity for
one-dimensional traces.

For the hidden-state trajectory, HaluNet uses a lightweight two-layer 1D
convolutional branch:
\begin{align}
Z^{(\mathrm{emb})}
&=\operatorname{ReLU}\!\left(
\operatorname{Conv1D}_{d_e\rightarrow d_h}
(X^{(\mathrm{emb})})
\right),\\
U^{(\mathrm{emb})}
&=\operatorname{ReLU}\!\left(
\operatorname{Conv1D}_{d_h\rightarrow d_h}(Z^{(\mathrm{emb})})
\right),\\
h^{(\mathrm{emb})}
&=\operatorname{Pool}(U^{(\mathrm{emb})})
\in\mathbb{R}^{d_h},
\end{align}
where $\operatorname{Pool}$ denotes adaptive average pooling over the
generated-token axis. Both convolutions use kernel size 3, stride 1, and
length-preserving padding. After permutation, the feature dimension is treated
as the channel dimension and the generated-token axis as the temporal dimension.
The branch output $h^{(\mathrm{emb})}$ summarizes local trajectory patterns
before pooling. We give the embedding branch sequence capacity because hidden
states form a high-dimensional semantic trajectory, where risk evidence may
appear as localized representational shifts in the generated sequence. The
branch encoder is intentionally shallow: the backbone model has already
processed the linguistic sequence, so HaluNet only needs to summarize
risk-related patterns over the extracted signals before reducing them to an
answer-level representation. Appendix~\ref{sec:token_attribution} provides
token-level Integrated Gradients~\cite{pmlr-v70-sundararajan17a} analysis illustrating how local signal patterns
contribute to the final risk score.

The main fusion module uses attention over the three branch representations.
Let
\begin{align}
h_r&\in\{h^{(\mathrm{ll})},h^{(\mathrm{ent})},h^{(\mathrm{emb})}\},\\
e_r&=v^\top\tanh(W_a h_r),\\
\alpha_r&=\frac{\exp(e_r)}{\sum_{r'}\exp(e_{r'})},\\
h_{\mathrm{fused}}
&=\sum_r \alpha_r h_r .
\end{align}
Attention fusion lets the detector adaptively emphasize the evidence source
that is most informative for a given answer while keeping the fusion module
small and branch-level. The output layer produces the risk score,
\begin{equation}
s_{\phi}(q,c,a)=
\sigma(w^{\top}h_{\mathrm{fused}}+b).
\end{equation}

The model is trained with binary cross-entropy on the weak labels:
\begin{equation}
\mathcal{L}=
-\frac{1}{N}\sum_i
\left[
y_i\log s_i+(1-y_i)\log(1-s_i)
\right].
\end{equation}

\subsection{Architecture Design Choices}

We evaluate two architectural choices: branch encoders and fusion modules.
For branch encoding, we compare global-average branches, CNN branches, and
HaluNet's mixed encoder, which averages scalar likelihood and entropy features
while encoding the hidden-state trajectory with a CNN. For fusion, we compare
attention fusion against concat-MLP fusion. As shown in
Section~\ref{sec:rq4_architecture}, the mixed encoder with attention fusion
achieves the strongest architecture-level results. We therefore use this
Emb-CNN + attention design as the main HaluNet architecture.

\section{Experiments}
\subsection{Experimental Setup}
Beyond the appendices explicitly referenced in the main text,
Appendices~\ref{sec:metric_details}--\ref{sec:human_annotation_protocol}
also provide full metric definitions, additional signal and layer analyses,
retrieval-augmented NQ experiments, robustness checks, prompt templates, and the annotation protocol.

\paragraph{Datasets and labels.}
We evaluate on three QA benchmarks: SQuAD 2.0 (SQuAD)~\cite{rajpurkar2018know},
TriviaQA~\cite{joshi2017triviaqa}, and NQ-Open (NQ)
~\cite{kwiatkowski2019natural}. For SQuAD and TriviaQA, we evaluate both
context-free and context-present settings, denoted by context ratio CR=0 and
CR=1, respectively. NQ is used as a context-free open-domain QA setting. Each
generated answer is labeled as supported or unsupported by the
LLM-as-a-Judge protocol described in
Section~\ref{sec:training_data_construction}. Table~\ref{tab:hallucination_summary}
summarizes dataset sizes and hallucination rates.

\begin{table}[t]
\centering
\small
\setlength{\tabcolsep}{3.2pt}
\renewcommand{\arraystretch}{1.05}
\caption{Dataset sizes and hallucination rates (\%) under different context
ratios. Hallucination rate is the percentage of generated answers labeled as
unsupported by the judge protocol.}
\label{tab:hallucination_summary}
\resizebox{\columnwidth}{!}{%
\begin{tabular}{llccc}
\toprule
\textbf{Dataset} & \textbf{Model}
& \textbf{Ori.}
& \textbf{Sampled}
& \textbf{Hallu.} \\
& & \textbf{Train/Test}
& \textbf{Train/Test}
& \textbf{Train/CR0/CR1} \\
\midrule
\textit{SQuAD} & Llama3-8B & 130k / 11.9k & 10k / 5.9k & 41.7 / 82.4 / 17.1 \\
\textit{SQuAD} & Qwen3-14B & 130k / 11.9k & 10k / 5.9k & 43.4 / 74.7 / 8.6 \\
\midrule
\textit{TriviaQA} & Llama3-8B & 12.3k / 3.1k & 10k / 3.1k & 19.6 / 29.8 / 10.1 \\
\textit{TriviaQA} & Qwen3-14B & 12.3k / 3.1k & 10k / 3.1k & 41.8 / 42.4 / 15.6 \\
\midrule
\textit{NQ} & Llama3-8B & 87.9k / 3.6k & 10k / 3.6k & 57.8 / 61.4 / -- \\
\textit{NQ} & Qwen3-14B & 87.9k / 3.6k & 10k / 3.6k & 71.1 / 71.6 / -- \\
\bottomrule
\end{tabular}
}
\end{table}

\paragraph{Baselines.}
We group baselines by their inference-time requirements.
\textit{Original-answer baselines} use only the original answer and signals
available from its generation, including \textbf{Predictive Entropy (PE)}
~\cite{farquhar2024detecting}, \textbf{Token Negative Log-Likelihood (T-NLL)}
~\cite{zhang2023enhancing}, and a supervised \textbf{Logistic} classifier
trained on the same feature set~\cite{hastie2009elements}.
\textit{Generation-time multi-sample baselines} require multiple sampled
generations to define the score, including \textbf{Embedding Variance (EmbVar)},
\textbf{Semantic Embedding Uncertainty (SEU)}~\cite{grewal2024improving}, and
\textbf{Semantic Entropy (SE)}~\cite{farquhar2024detecting}.
\textit{Evaluation-time baselines} require additional sampled answers or LLM
calls after the original answer is produced, including \textbf{SelfCheckGPT}
~\cite{manakul2023selfcheckgpt} and prediction-based \(\bm{P(\mathrm{True})}\)~\cite{farquhar2024detecting}.

\paragraph{Metrics.}
All paper-facing experiments use the same four metrics. AUROC and AUPRC measure
whether risk scores rank unsupported answers above supported answers, following
standard threshold-free evaluation for binary detection and imbalanced settings
~\cite{davis2006relationship,saito2015precision}. RetAcc@50 follows selective
prediction evaluation~\cite{geifman2017selective}: it rejects the highest-risk
50\% of answers and reports the accuracy of the remaining low-risk half.
Lift@20 follows top-ranked retrieval evaluation: it measures the error rate
among the top 20\% highest-risk answers relative to the overall error rate,
summarizing how well a detector concentrates errors at the top of the risk
ranking~\cite{manning2008introduction}.

\begin{table*}[t]
\centering
\footnotesize
\setlength{\tabcolsep}{2.6pt}
\renewcommand{\arraystretch}{1.05}
\caption{Main results under CR=1. All scores are answer-level hallucination risk metrics; higher is better. Best and second-best results within each backbone--dataset block are bolded and underlined. Latency is reported per 100 examples in seconds; cell shading is a visual cue for latency magnitude.}
\label{tab:main-cr1}
\resizebox{\textwidth}{!}{%
\begin{tabular}{ll|cccc|cccc|rr}
\toprule
\multicolumn{2}{c|}{\multirow{2}{*}{\textbf{Backbone / Method}}}  & \multicolumn{4}{c|}{\textbf{SQuAD}} & \multicolumn{4}{c|}{\textbf{TriviaQA}} & \multicolumn{2}{c}{\textbf{Latency /100}} \\
\cmidrule(lr){3-6}\cmidrule(lr){7-10}\cmidrule(lr){11-12}
\multicolumn{2}{c|}{}  & AUROC & AUPRC & RetAcc@50 & Lift@20 & AUROC & AUPRC & RetAcc@50 & Lift@20  & \multicolumn{1}{c}{Sample} & \multicolumn{1}{c}{Infer} \\
\midrule
\multirow{9}{*}{\rotatebox{90}{\textbf{Llama3-8B}}} & PE & 0.732 & 0.307 & 0.933 & 1.987 & 0.723 & 0.247 & 0.955 & 2.344 & \cellcolor{red!20}105.435 & \cellcolor{green!10}0.001 \\
 & T-NLL & 0.628 & 0.243 & 0.874 & 2.058 & 0.571 & 0.124 & 0.914 & 1.608 & \cellcolor{red!10}80.780 & \cellcolor{green!10}0.001 \\
 & EmbVar & 0.756 & 0.380 & 0.942 & 2.305 & 0.753 & 0.316 & 0.962 & 2.717 & \cellcolor{red!20}727.163 & \cellcolor{green!10}0.109 \\
 & SE & 0.620 & 0.252 & 0.869 & 1.937 & 0.735 & 0.252 & 0.950 & 2.942 & \cellcolor{red!20}743.258 & \cellcolor{red!10}29.927 \\
 & SEU & 0.764 & 0.416 & 0.940 & 2.409 & 0.792 & 0.406 & 0.965 & 3.167 & \cellcolor{red!20}731.105 & \cellcolor{yellow!30}2.068 \\
 & SelfCheckGPT & \underline{0.772} & 0.364 & \underline{0.944} & 2.424 & 0.779 & 0.228 & 0.978 & 2.674 & \cellcolor{red!20}760.002 & \cellcolor{red!10}87.161 \\
 & $P(\mathrm{True})$ & 0.673 & 0.343 & 0.895 & 2.012 & 0.774 & 0.314 & 0.960 & 2.781 & \cellcolor{red!10}81.557 & \cellcolor{red!10}30.413 \\
 & Logistic & 0.748 & \underline{0.462} & 0.925 & \underline{2.609} & \underline{0.854} & \underline{0.553} & \underline{0.979} & \underline{3.569} & \cellcolor{red!10}98.849 & \cellcolor{green!10}0.028 \\
 & \textbf{HaluNet} & \textbf{0.839} & \textbf{0.562} & \textbf{0.965} & \textbf{2.866} & \textbf{0.893} & \textbf{0.637} & \textbf{0.984} & \textbf{3.859} & \cellcolor{red!10}97.663 & \cellcolor{green!10}0.124 \\
\midrule
\multirow{9}{*}{\rotatebox{90}{\textbf{Qwen3-14B}}} & PE & 0.705 & 0.167 & 0.960 & 2.310 & 0.719 & 0.311 & 0.925 & 2.175 & \cellcolor{red!20}127.035 & \cellcolor{green!10}0.001 \\
 & T-NLL & 0.558 & 0.112 & 0.924 & 1.464 & 0.554 & 0.174 & 0.861 & 1.427 & \cellcolor{red!20}131.440 & \cellcolor{green!10}0.001 \\
 & EmbVar & 0.724 & 0.240 & 0.960 & 2.705 & 0.649 & 0.264 & 0.909 & 1.737 & \cellcolor{red!20}1289.101 & \cellcolor{green!10}0.126 \\
 & SE & 0.661 & 0.165 & 0.943 & 2.345 & 0.652 & 0.243 & 0.903 & 1.904 & \cellcolor{red!20}1253.741 & \cellcolor{red!10}24.100 \\
 & SEU & \underline{0.742} & \underline{0.308} & 0.960 & \underline{2.793} & 0.658 & 0.274 & 0.910 & 1.708 & \cellcolor{red!20}1302.565 & \cellcolor{yellow!30}1.948 \\
 & SelfCheckGPT & 0.652 & 0.203 & \underline{0.966} & 2.205 & 0.776 & 0.567 & 0.910 & 2.663 & \cellcolor{red!20}1267.408 & \cellcolor{red!10}74.015 \\
 & $P(\mathrm{True})$ & 0.677 & 0.225 & 0.950 & 2.148 & 0.746 & 0.392 & 0.932 & 2.396 & \cellcolor{red!20}118.871 & \cellcolor{red!10}53.593 \\
 & Logistic & 0.707 & 0.249 & 0.955 & 2.559 & \underline{0.796} & \underline{0.594} & \underline{0.936} & \underline{3.042} & \cellcolor{red!20}121.403 & \cellcolor{green!10}0.042 \\
 & \textbf{HaluNet} & \textbf{0.781} & \textbf{0.364} & \textbf{0.972} & \textbf{2.861} & \textbf{0.856} & \textbf{0.698} & \textbf{0.961} & \textbf{3.417} & \cellcolor{red!20}128.006 & \cellcolor{green!10}0.119 \\
\bottomrule
\end{tabular}
}
\end{table*}

\paragraph{Implementation.}
We evaluate HaluNet with two backbone models,
LLM-Research/Meta-Llama-3-8B-Instruct~\cite{grattafiori2024llama} and Qwen/Qwen3-14B~\cite{yang2025qwen3}.
HaluNet is trained with AdamW using learning rate $10^{-4}$, batch size 32, and
early stopping on validation AUROC. Answer generation and signal extraction are
run on 8 NVIDIA RTX 3090 GPUs. The original LLM-as-a-Judge labels are produced
by the answer-generating backbone via three repeated judge rounds and
majority voting. For independent validation, we use Qwen/Qwen3.5-35B-A3B\footnote{\url{https://huggingface.co/Qwen/Qwen3.5-35B-A3B}} in text-only
mode as a separate judge, aggregating three complementary prompts by majority
vote.

\subsection{RQ1: Main Performance and OOD Transfer}
\label{sec:rq1_main}

RQ1 evaluates whether HaluNet can rank answer-level hallucination risk using
only single-generation internal signals. Table~\ref{tab:main-cr1} reports the
main context-present results on SQuAD and TriviaQA (CR=1). HaluNet achieves the
strongest overall performance among probability-based scores, latent-space
uncertainty estimators, sampling-based detectors, and LLM-based self-evaluation
baselines. On Llama3-8B, it obtains 0.839/0.893 AUROC on SQuAD/TriviaQA,
improving over the strongest non-HaluNet baselines by 0.067 and 0.039. On
Qwen3-14B, HaluNet also ranks first across AUROC, AUPRC, RetAcc@50, and
Lift@20 on both datasets. The larger gains in AUPRC and Lift@20 indicate that
HaluNet concentrates unsupported answers near the top of the risk ranking, beyond
improvements in global separability alone. These results suggest that HaluNet
reduces the accuracy--cost trade-off: it uses only the internal signals from the
original generation, yet outperforms methods that require additional samples or
LLM calls. Full CR=0 results are provided in
Appendix~\ref{sec:appendix_main_cr0}.

\begin{figure}[t]
\centering
\includegraphics[width=\columnwidth]{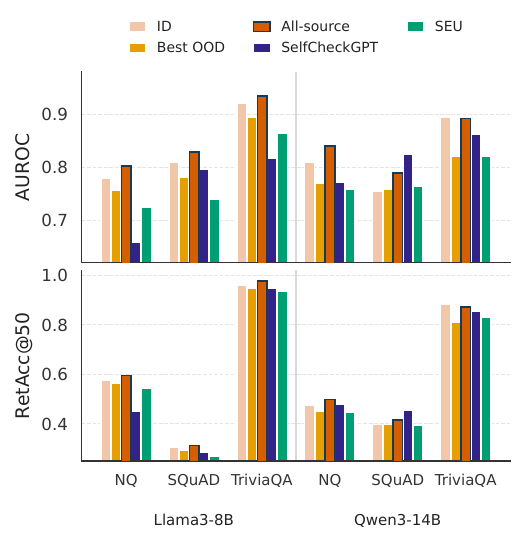}
\caption{OOD transfer under CR=0.}
\label{fig:ood-transfer-summary}
\end{figure}

We further evaluate cross-dataset transfer under CR=0, comparing target-dataset
training (ID), the best non-target source (Best OOD), all-source training,
SelfCheckGPT and SEU. As shown in Fig.~\ref{fig:ood-transfer-summary},
all-source HaluNet outperforms ID on all three Llama3-8B targets in both AUROC
and RetAcc@50; on Qwen3-14B, it improves NQ and SQuAD, and remains nearly tied
with ID on TriviaQA in AUROC, with slightly lower RetAcc@50. External references
are setting-dependent: SelfCheckGPT is competitive on Qwen3-14B SQuAD, but
all-source HaluNet shows the strongest overall transfer pattern, while SEU
remains consistently below it. The full single-source results in
Appendix~\ref{sec:appendix_ood} further reveal asymmetric transfer, e.g.,
TriviaQA transfers reasonably to Llama3-8B SQuAD but poorly to Qwen3-14B SQuAD.
Overall, multi-source supervision helps learn reusable hallucination-risk
patterns, although source--target match still affects transfer.

\subsection{RQ2: Human-Annotated Risk Ranking}
\label{sec:rq2_human_ranking}

RQ2 evaluates whether HaluNet's risk score aligns with independent human
judgments, not merely with judge-generated supervision. We construct a
300-example human evaluation set balanced along two dimensions: across datasets
(NQ, SQuAD, and TriviaQA) and, within each source split, across the original
judge labels. The set covers both context-free and context-present settings for
SQuAD and TriviaQA, while NQ is evaluated under CR=0. Three trained annotators
independently annotate each example, and the final binary label is obtained by
majority vote. After excluding 16 examples marked ambiguous, 284 valid examples
remain. Table~\ref{tab:human-eval-baseline-risk} compares HaluNet with baselines
rerun under the same human-label protocol.

\begin{table}[t]
\centering
\footnotesize
\setlength{\tabcolsep}{3.4pt}
\renewcommand{\arraystretch}{1.05}
\caption{Human-annotated answer-level risk ranking on 284 valid examples.
Gen. ($K=9$) counts sampled generations beyond the original answer; LLM counts
additional test-time LLM or judge calls.}
\label{tab:human-eval-baseline-risk}
\resizebox{\columnwidth}{!}{%
\begin{tabular}{lccrrrr}
\toprule
\textbf{Method} & \textbf{Gen.} & \textbf{LLM} & \textbf{AUROC} & \textbf{AUPRC} & \textbf{RetAcc@50} & \textbf{Lift@20} \\
\midrule
T-NLL & 0 & 0 & 0.598 & 0.567 & 0.585 & 1.386 \\
PE & 0 & 0 & 0.737 & 0.664 & 0.704 & 1.536 \\
EmbVar & K & 0 & 0.785 & 0.763 & 0.739 & 1.686 \\
SEU & K & 0 & 0.770 & 0.727 & 0.746 & 1.573 \\
SE & K & 0 & 0.757 & 0.648 & 0.754 & 1.611 \\
$P(\mathrm{True})$ & 0 & 1 & 0.753 & 0.731 & 0.725 & 1.723 \\
SelfCheckGPT & K & 0 & 0.769 & 0.720 & 0.732 & 1.611 \\
Logistic & 0 & 0 & \underline{0.835} & \underline{0.834} & \underline{0.761} & \underline{1.911} \\
\textbf{HaluNet} & 0 & 0 & \textbf{0.874} & \textbf{0.869} & \textbf{0.824} & \textbf{2.060} \\
\bottomrule
\end{tabular}

}
\end{table}

HaluNet achieves the strongest human-aligned ranking, with 0.874 AUROC, 0.869
AUPRC, 0.824 RetAcc@50, and 2.06 Lift@20. Compared with SEU, a strong non-LLM uncertainty baseline, and SelfCheckGPT, a representative sampling-based baseline, HaluNet separates human-confirmed errors more sharply.
This supports the central claim that internal-signal risk patterns learned from
judge-generated supervision remain meaningful under independent human labels.

\begin{figure}[t]
\centering
\includegraphics[width=\columnwidth]{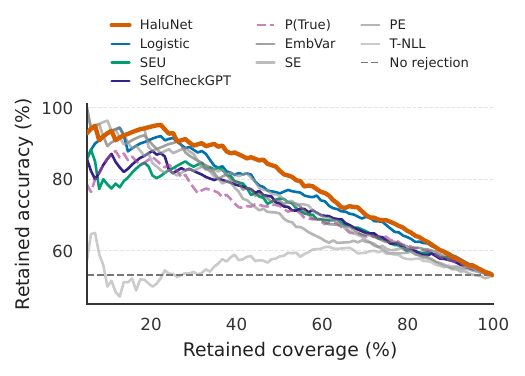}
\caption{Human-annotated risk-coverage curves. Higher retained accuracy at lower
coverage indicates better ranking after rejecting high-risk answers.}
\label{fig:human-eval-risk-coverage}
\end{figure}

Fig.~\ref{fig:human-eval-risk-coverage} further shows that HaluNet maintains the
highest retained accuracy over most of the coverage range. This indicates that
its score is effective for selective QA, where high-risk answers can be routed
to verification, regeneration, or abstention.

\subsection{RQ3: Signal Complementarity}
\label{sec:rq3_signals}

RQ3 examines whether the three internal signal families provide complementary
evidence for answer-level risk estimation. We use the CR=0 ID ablation setting
and restrict HaluNet to different signal subsets. For each dataset--backbone
pair, we report AUROC for each signal set using its branch encoding variant. To
make the comparison focus on the signal families, all variants use the same
attention-based fusion module.

\begin{figure}[t]
\centering
\includegraphics[width=\columnwidth]{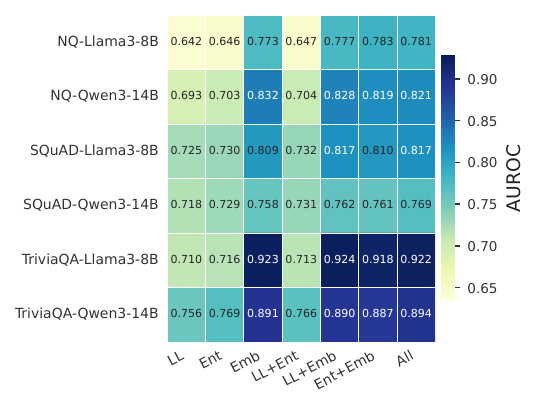}
\caption{Signal complementarity under CR=0.}
\label{fig:signal-complementarity}
\end{figure}

Fig.~\ref{fig:signal-complementarity} shows that likelihood and entropy alone
provide useful but limited evidence, while embedding-based variants are
consistently stronger. Adding likelihood or entropy to embeddings often improves
over using embeddings alone. These results support the core design motivation of
HaluNet: hallucination risk is not fully captured by a single internal statistic,
and probabilistic, distributional, and representational signals provide partially
distinct evidence for answer-level risk estimation.

\subsection{RQ4: Architecture Choice}
\label{sec:rq4_architecture}

RQ4 studies how branch encoding and fusion design affect HaluNet. We compare
three branch encoders with two lightweight fusion modules, attention fusion and
concat-MLP fusion, using SQuAD CR=0 as the main stress setting.

\begin{figure}[t]
\centering
\includegraphics[width=\columnwidth]{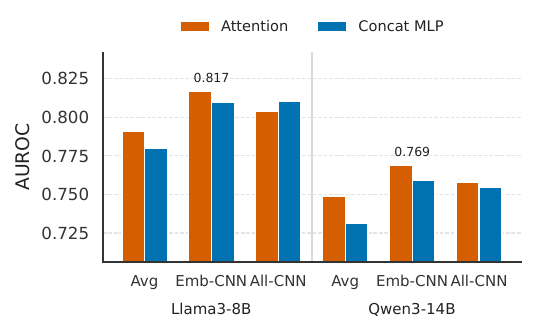}
\caption{Architecture ablation on SQuAD under CR=0.}
\label{fig:architecture-ablation}
\end{figure}

Fig.~\ref{fig:architecture-ablation} shows that architecture choices affect
HaluNet in two ways. First, the mixed encoder outperforms the all-average
encoder, indicating that the high-dimensional hidden-state trajectory benefits
from sequence-aware encoding. Second, attention fusion gives the strongest
overall results in the mixed encoder, although concat-MLP remains competitive in
some settings. We therefore use the Emb-CNN + attention configuration as the
main HaluNet architecture. These results suggest that signal-appropriate branch
encoding is more important than heavy fusion, and that lightweight
branch-level fusion is sufficient to combine the three internal views.

\subsection{RQ5: Reliability Validation and Error Analysis}
\label{sec:rq5_label_reliability}

RQ5 examines whether HaluNet's learned risk scores and judge-generated
supervision remain reliable beyond a single judge source. We evaluate
dataset-wise alignment between HaluNet scores and human labels, and label
agreement among the original judge, an independent Qwen3.5 multi-judge protocol,
and human annotations.

\begin{table}[t]
\centering
\small
\setlength{\tabcolsep}{3.5pt}
\renewcommand{\arraystretch}{1.05}
\caption{Dataset-wise HaluNet risk validation against human labels on the
human evaluation set. Err@20 is the human-confirmed error rate (\%) among the top
20\% highest-risk answers. Ret@50 denotes RetAcc@50.}
\label{tab:human-risk-by-dataset}
\resizebox{\columnwidth}{!}{%
\begin{tabular}{lrrrrrr}
\toprule
\textbf{Subset} & \textbf{$n$} & \textbf{AUROC} & \textbf{AUPRC} & \textbf{Ret@50} & \textbf{Err@20} & \textbf{Lift@20} \\
\midrule
Overall & 284 & 0.874 & 0.869 & 0.824 & 96.5 & 2.060 \\
NQ & 97 & 0.846 & 0.859 & 0.784 & 95.0 & 1.961 \\
SQuAD & 93 & 0.884 & 0.833 & 0.882 & 89.5 & 2.134 \\
TriviaQA & 94 & 0.932 & 0.949 & 0.872 & 100.0 & 2.000 \\
\bottomrule
\end{tabular}
}
\end{table}

Table~\ref{tab:human-risk-by-dataset} shows that HaluNet's human-aligned risk
ranking is not driven by a single dataset: Lift@20 remains close to or above
$2\times$ across NQ, SQuAD, and TriviaQA, with high AUROC on all three subsets.

\begin{table}[t]
\centering
\small
\setlength{\tabcolsep}{3.5pt}
\renewcommand{\arraystretch}{1.05}
\caption{Human label reliability. Agreement and Cohen's $\kappa$ compare
binary labels from the original judge, an independent Qwen3.5 multi-judge
protocol, and human annotations.}
\label{tab:judge-agreement}
\resizebox{\columnwidth}{!}{%
\begin{tabular}{lrrr}
\toprule
\textbf{Comparison} & \textbf{$n$} & \textbf{Agreement} & \textbf{$\kappa$} \\
\midrule
Original judge vs. human & 284 & 0.940 & 0.880 \\
Qwen3.5 multi-judge vs. human & 284 & 0.919 & 0.836 \\
Original judge vs. Qwen3.5 & 300 & 0.857 & 0.714 \\
\bottomrule
\end{tabular}

}
\end{table}

Table~\ref{tab:judge-agreement} reports high agreement with human annotations
for both the original judge (0.940 agreement, $\kappa=0.880$) and the independent
Qwen3.5 multi-judge protocol (0.919 agreement, $\kappa=0.836$). These results
support using judge-generated labels as scalable weak supervision, with human
annotations serving as external validation.

Human annotations also provide a compact error analysis. Errors are more
frequent under CR=0 than CR=1 (50.5\% vs. 39.1\%), and 45 of the 57 top-20\%
HaluNet-risk examples come from CR=0, suggesting that many high-risk cases occur
without supporting context. CR=1 still contains confirmed errors, indicating
remaining context-use failures. Among examples with error-type notes, entity and numeric/date errors are most
frequent. These sparse annotations provide qualitative evidence and are not
intended to constitute a complete error taxonomy.

\section{Conclusion}
This work shows that answer-level hallucination risk can be estimated directly
from the generation process. HaluNet combines likelihood, entropy, and
hidden-state signals into a single risk score, allowing white-box QA systems to
rank hallucination risk without extra sampling or external verification.

Our results suggest a scalable path for reliability modeling: judge-generated
labels can weakly supervise lightweight risk estimators. At deployment, HaluNet
can act as an early risk filter, routing high-risk answers to verification,
regeneration, or abstention. Future work should extend this approach to
long-form and multi-turn QA, and to settings with more limited model access.

\section*{Limitations}
This work focuses on answer-level hallucination detection for QA. HaluNet is
designed to rank generated answers by reliability, rather than to provide a
fine-grained diagnosis of all possible error sources. Future work could combine
internal-signal risk estimation with retrieval analysis, attribution methods,
or reasoning analysis to better characterize different types of answer errors.

Our experiments cover three QA benchmarks, two backbone LLMs, and both
in-domain and out-of-domain settings. Broader evaluation across more model
families, model scales, instruction-tuned variants, and domain-specific QA
systems would further clarify how well internal risk patterns transfer across
deployment conditions.

Finally, HaluNet is evaluated mainly on single-turn QA with relatively short
answers. Extending the framework to long-form, multi-turn, and tool-augmented QA
would require modeling longer reliability trajectories and aggregating risk
across answer segments or dialogue turns. Larger and more diverse human
evaluations would also further strengthen reliability validation.

\section*{Ethical Considerations}

\paragraph{Data sources.}
This work uses publicly available QA benchmarks, including SQuAD, TriviaQA, and
Natural Questions. We use benchmark questions, reference answers, and contexts
when available, and generate model answers for research evaluation. The study
does not use private user conversations, personal records, or sensitive
application data.

\paragraph{Human annotation.}
We conduct a small human evaluation to validate hallucination-risk scores and
judge-generated labels. Annotators received written guidelines and a calibration
set before annotation. They judged whether generated answers were supported by
the provided question, reference answer, and available context when present.
Annotators could mark examples as ambiguous, and such examples were excluded
from human-label metric computation. Annotators were informed that their labels
would be used for academic research, agreed to participate before the task
began, and were compensated for their time. We do not report annotator identities
or personal information.

\paragraph{Privacy and sensitive content.}
The annotation task uses benchmark QA examples and model-generated answers. It
does not require annotators to provide personal information or evaluate
sensitive user data. Although benchmark datasets may contain naturally occurring
names or factual references, we use them only for standard research evaluation
and do not attempt to infer private information about individuals.

\paragraph{Risks and misuse.}
HaluNet produces an answer-level hallucination-risk score, not a guarantee of
truthfulness or a complete explanation of model behavior. A possible misuse is
to treat low-risk outputs as verified facts or to use the detector as a
replacement for human review in high-stakes domains such as medical, legal, or
financial decision making. The intended use is to support research on
hallucination detection and to help route high-risk answers to verification,
regeneration, or abstention. HaluNet should therefore be used as an auxiliary
reliability signal rather than as a standalone factuality oracle.

\section*{Acknowledgments}
We thank the open-source communities whose models, datasets, and software libraries made this work possible. We are also grateful to Jiayang Gao and Nannan Sun for supporting this project through research funding.

\bibliography{my}

\clearpage
\appendix

\section{Metric Details}
\label{sec:metric_details}

All risk scores are oriented so that larger values indicate higher
hallucination risk. Let $y_i\in\{0,1\}$ be the evaluation label for answer
$i$, where $1$ denotes an unsupported or hallucinated answer.

\paragraph{AUROC and AUPRC.}
AUROC and AUPRC evaluate whether unsupported answers are ranked above supported
answers. AUPRC is especially informative when the unsupported-answer rate is
imbalanced.

\paragraph{RetAcc@50.}
RetAcc@50 measures retained-answer accuracy after rejecting the highest-risk
half of the answers. If $\mathcal{R}_{50}$ denotes the 50\% of answers with the
lowest risk scores, then
\[
\mathrm{RetAcc@50}
=
\frac{1}{|\mathcal{R}_{50}|}
\sum_{i\in\mathcal{R}_{50}}\mathbb{I}[y_i=0].
\]
This metric follows the selective prediction perspective: a useful risk score
should leave a more reliable retained set after high-risk answers are rejected.

\paragraph{Lift@20.}
Lift@20 measures error concentration in the highest-risk region:
\[
\mathrm{Lift@20}
=
\frac{\mathrm{ErrRate}(\mathrm{Top}\;20\%\;\mathrm{risk})}
{\mathrm{ErrRate}(\mathrm{all})}.
\]
A value above one indicates that unsupported answers are concentrated among the
top-risk examples.

\section{Internal Signal Correlation}
\label{sec:signal_correlation}

\begin{figure}[t]
\centering
\includegraphics[width=\columnwidth]{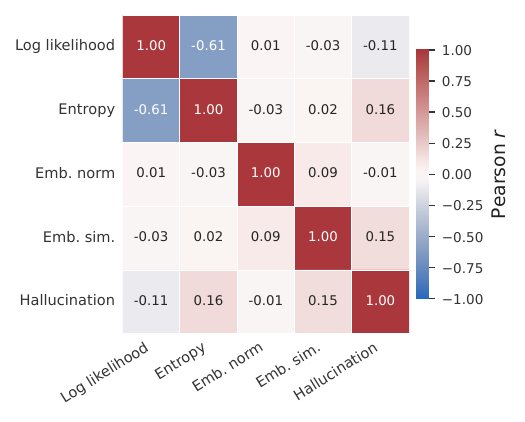}
\caption{Internal-signal correlations on SQuAD CR=0 with Llama3-8B.}
\label{fig:internal-signal-correlation}
\end{figure}

Fig.~\ref{fig:internal-signal-correlation} provides an empirical check of how
simple internal-signal summaries relate to the hallucination label. We compute
answer-level summaries on SQuAD CR=0 with Llama3-8B, including log-likelihood,
predictive-entropy, and hidden-state trajectory statistics. The correlations
show that these summaries are related but not interchangeable, supporting the
use of multiple internal views rather than a single one.
The strongest correlations are not concentrated in one feature family, which is
consistent with the main method design: likelihood, entropy, and hidden-state
summaries provide overlapping but distinct evidence.

\section{Token-Level Attribution}
\label{sec:token_attribution}

We use Integrated Gradients~\cite{pmlr-v70-sundararajan17a} to qualitatively inspect which token-level signals
contribute to the risk logit. Positive values increase the predicted risk, while
negative values decrease it. For readability, each signal row is normalized
independently in the heatmaps; this visualization should not be interpreted as
comparing absolute importance across signal families. The examples below are
intended as qualitative inspections of HaluNet's learned score, not as causal
explanations of the backbone model's generation behavior.

\begin{figure}[t]
\centering
\includegraphics[width=\columnwidth]{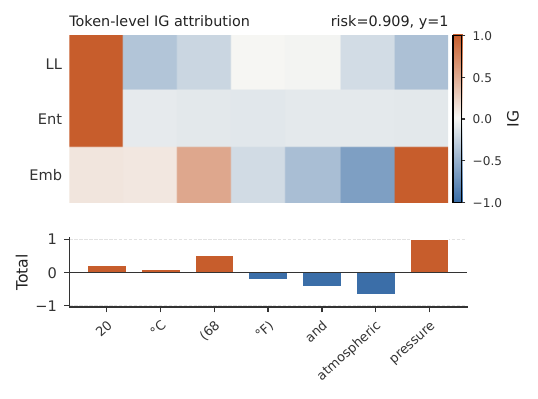}
\caption{Token attribution: wrong condition phrase.}
\label{fig:token-attr-oxygen}
\end{figure}

Fig.~\ref{fig:token-attr-oxygen} shows a context-adjacent mismatch. The question
asks the state in which oxygen is shipped in bulk, while the context states
that oxygen is transported ``as a liquid.'' The generated answer instead
selects a nearby condition phrase, ``20 $^\circ$C (68 $^\circ$F) and
atmospheric pressure.'' HaluNet assigns high risk ($s_\phi=0.909$), with strong
local contribution around the mismatched phrase.

\begin{figure}[t]
\centering
\includegraphics[width=\columnwidth]{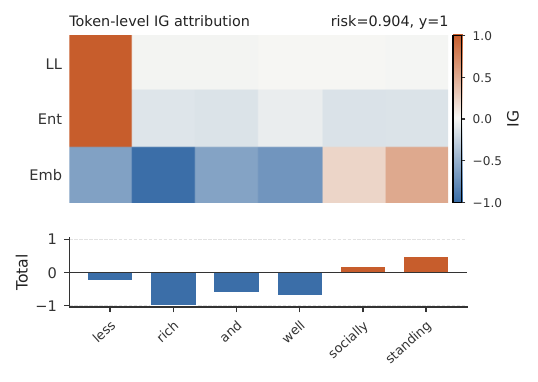}
\caption{Token attribution: semantic mismatch.}
\label{fig:token-attr-mongols}
\end{figure}

Fig.~\ref{fig:token-attr-mongols} shows an answer mismatch in which the
reference describes Mongols who lived in poverty and were ill treated, whereas
the generated answer says ``less rich and well socially standing.'' The
attribution pattern highlights local risk evidence around the generated
answer span.

\begin{figure}[t]
\centering
\includegraphics[width=\columnwidth]{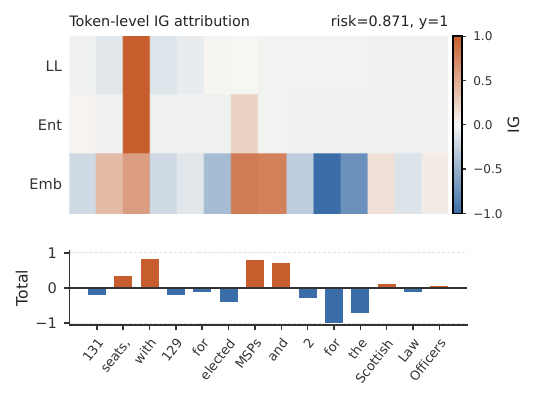}
\caption{Token attribution: answer-type mismatch.}
\label{fig:token-attr-chamber}
\end{figure}

Fig.~\ref{fig:token-attr-chamber} shows an answer-type mismatch. The question
asks for the seating arrangement of the debating chamber, whose reference
answer is ``hemicycle.'' The generated answer instead gives seat counts and
office-holder seats. HaluNet assigns high risk ($s_\phi=0.871$), and the
attribution pattern marks the answer region as risk-relevant.

\section{Layer Sensitivity}
\label{sec:layer_sensitivity}

\begin{figure}[t]
\centering
\includegraphics[width=\columnwidth]{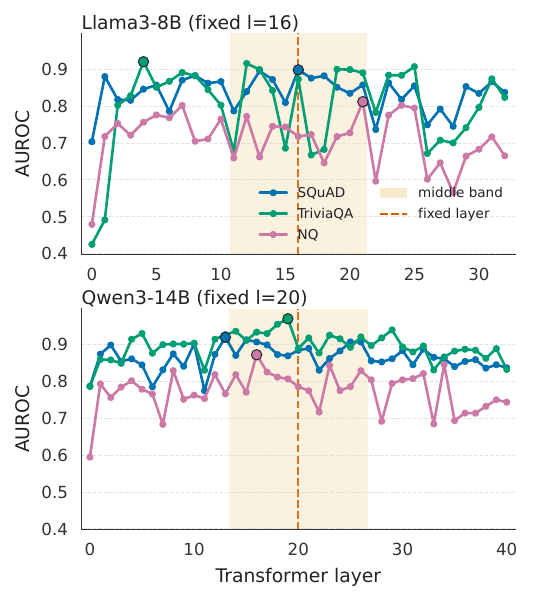}
\caption{Hidden-layer sensitivity.}
\label{fig:layer-contribution}
\end{figure}

We use the layer sweep as an analysis of where hallucination-risk information
appears across the backbone, with separate comparisons for early, middle, and
late transformer representations. The fixed layer used in the main model is
pre-specified and is not selected using test or human-evaluation labels.
Fig.~\ref{fig:layer-contribution} reports per-layer AUROC, and
Table~\ref{tab:layer-band} groups layers into early, middle, and late thirds.
The results support using a fixed intermediate representation in the main
experiments.
The layer sweep does not identify one universally dominant layer. Instead, the
middle band is broadly competitive, so the main model uses a pre-specified
intermediate layer as a cost-conscious and stable feature choice rather than as
a tuned optimum.

\begin{table}[t]
\centering
\small
\caption{Layer-band sensitivity of HaluNet. We group transformer layers into early, middle, and late thirds and report mean AUROC within each band. This analysis evaluates where hallucination-relevant representations tend to emerge, rather than treating a single layer as universally optimal.}
\label{tab:layer-band}
\resizebox{\columnwidth}{!}{%
\begin{tabular}{llccc}
\toprule
\textbf{Dataset} & \textbf{Backbone} & \textbf{Early} & \textbf{Middle} & \textbf{Late} \\
\midrule
SQuAD & Llama3-8B & 0.836 & \textbf{0.855} & 0.814 \\
SQuAD & Qwen3-14B & 0.847 & \textbf{0.887} & 0.856 \\
TriviaQA & Llama3-8B & 0.783 & \textbf{0.812} & 0.798 \\
TriviaQA & Qwen3-14B & 0.879 & \textbf{0.922} & 0.883 \\
NQ & Llama3-8B & \textbf{0.723} & 0.721 & 0.683 \\
NQ & Qwen3-14B & 0.760 & \textbf{0.799} & 0.758 \\
\bottomrule
\end{tabular}
}
\end{table}

\section{CR=0 Main Results}
\label{sec:appendix_main_cr0}

Table~\ref{tab:main-cr0-appendix} reports the full context-free main results
using the same answer-level metrics as the main text. NQ is evaluated only under
CR=0 because it is treated as an open-domain QA setting.
HaluNet is strongest or near strongest across most CR=0 blocks, especially on
Llama3-8B where it leads all four metrics for SQuAD and TriviaQA and leads the
main ranking metrics for NQ. Qwen3-14B shows a more competitive low-context
setting: SelfCheckGPT is strong on SQuAD, but HaluNet remains best on NQ AUROC
and TriviaQA AUROC/AUPRC, supporting the broader conclusion that single-trace
internal signals provide a robust low-cost risk estimate.

\begin{table*}[t]
\centering
\scriptsize
\setlength{\tabcolsep}{2.6pt}
\caption{Main results under CR=0. All scores use the unified answer-centric metrics.}
\label{tab:main-cr0-appendix}
\resizebox{\textwidth}{!}{%
\begin{tabular}{ll|rrrr|rrrr|rrrr}
\toprule
\textbf{Backbone} & \textbf{Method} & \multicolumn{4}{c|}{\textbf{NQ}} & \multicolumn{4}{c|}{\textbf{SQuAD}} & \multicolumn{4}{c}{\textbf{TriviaQA}} \\
\cmidrule(lr){3-6}\cmidrule(lr){7-10}\cmidrule(lr){11-14}
 & & AUROC & AUPRC & RetAcc@50 & Lift@20 & AUROC & AUPRC & RetAcc@50 & Lift@20 & AUROC & AUPRC & RetAcc@50 & Lift@20 \\
\midrule
\multirow{9}{*}{\rotatebox{90}{\textbf{Llama3-8B}}} & PE & 0.663 & 0.736 & 0.484 & 1.267 & 0.710 & 0.898 & 0.256 & 1.105 & 0.711 & 0.479 & 0.834 & 1.765 \\
 & T-NLL & 0.593 & 0.678 & 0.451 & 1.215 & 0.641 & 0.871 & 0.252 & 1.115 & 0.622 & 0.399 & 0.767 & 1.754 \\
 & EmbVar & 0.664 & 0.732 & 0.497 & 1.267 & 0.722 & 0.912 & 0.266 & 1.135 & 0.826 & 0.694 & 0.912 & 2.502 \\
 & SE & 0.683 & 0.710 & 0.527 & 1.242 & 0.730 & 0.902 & 0.269 & 1.138 & 0.835 & 0.644 & 0.924 & 2.469 \\
 & SEU & 0.724 & 0.800 & 0.541 & 1.409 & 0.739 & 0.913 & 0.267 & 1.134 & 0.864 & 0.739 & 0.933 & 2.600 \\
 & SelfCheckGPT & 0.659 & 0.727 & 0.452 & 1.150 & 0.797 & 0.926 & 0.284 & 1.131 & 0.816 & 0.546 & 0.946 & 1.978 \\
 & $P(\mathrm{True})$ & 0.692 & 0.765 & 0.524 & 1.335 & 0.659 & 0.897 & 0.244 & 1.129 & 0.835 & 0.735 & 0.903 & 2.573 \\
 & Logistic & \underline{0.775} & \underline{0.829} & \textbf{0.584} & \underline{1.465} & \underline{0.804} & \underline{0.946} & \underline{0.303} & \underline{1.184} & \underline{0.900} & \underline{0.813} & \underline{0.950} & \underline{2.842} \\
 & \textbf{HaluNet} & \textbf{0.779} & \textbf{0.851} & \underline{0.575} & \textbf{1.501} & \textbf{0.810} & \textbf{0.947} & \textbf{0.304} & \textbf{1.187} & \textbf{0.922} & \textbf{0.862} & \textbf{0.960} & \textbf{2.977} \\
\midrule
\multirow{9}{*}{\rotatebox{90}{\textbf{Qwen3-14B}}} & PE & 0.700 & 0.838 & 0.402 & 1.222 & 0.706 & 0.851 & 0.364 & 1.169 & 0.787 & 0.704 & 0.796 & 1.752 \\
 & T-NLL & 0.619 & 0.779 & 0.355 & 1.230 & 0.642 & 0.807 & 0.351 & 1.157 & 0.601 & 0.511 & 0.634 & 1.547 \\
 & EmbVar & 0.714 & 0.854 & 0.406 & 1.269 & 0.711 & 0.860 & 0.363 & 1.214 & 0.797 & 0.753 & 0.811 & 1.962 \\
 & SE & 0.758 & 0.845 & 0.449 & 1.275 & 0.758 & 0.873 & 0.400 & 1.250 & 0.791 & 0.689 & 0.814 & 1.925 \\
 & SEU & 0.758 & 0.882 & 0.444 & 1.309 & \underline{0.765} & 0.893 & 0.393 & 1.251 & 0.820 & 0.793 & 0.830 & 2.063 \\
 & SelfCheckGPT & 0.771 & 0.874 & \textbf{0.479} & \underline{1.353} & \textbf{0.825} & \textbf{0.924} & \textbf{0.452} & \textbf{1.323} & \underline{0.863} & \underline{0.844} & \underline{0.855} & \underline{2.259} \\
 & $P(\mathrm{True})$ & 0.749 & 0.868 & 0.437 & 1.288 & 0.662 & 0.843 & 0.343 & 1.189 & 0.846 & 0.838 & 0.836 & 2.192 \\
 & Logistic & \underline{0.804} & \underline{0.909} & \underline{0.475} & \textbf{1.354} & 0.753 & \underline{0.898} & \underline{0.400} & \underline{1.277} & 0.846 & 0.834 & 0.832 & 2.196 \\
 & \textbf{HaluNet} & \textbf{0.810} & \textbf{0.912} & 0.473 & \textbf{1.354} & 0.755 & 0.896 & 0.396 & 1.266 & \textbf{0.894} & \textbf{0.891} & \textbf{0.883} & \textbf{2.292} \\
\bottomrule
\end{tabular}
}
\end{table*}

\section{OOD Transfer Details}
\label{sec:appendix_ood}

Table~\ref{tab:ood-gain-appendix} provides the complete CR=0 source--target
transfer matrix for HaluNet. In addition to ID and all-source training, it
reports each single-source transfer direction, e.g., SQuAD$\rightarrow$NQ and
NQ$\rightarrow$SQuAD. This full matrix is important because transfer is not
symmetric: the best non-target source varies by target dataset and backbone.
All-source training is usually the most stable choice, improving over ID on
Llama3-8B and on two of three Qwen3-14B targets.

\begin{table*}[t]
\centering
\footnotesize
\setlength{\tabcolsep}{4pt}
\renewcommand{\arraystretch}{1.04}
\caption{Complete OOD transfer results under CR=0. Each row reports HaluNet
trained on the source shown in the third column and evaluated on the target
dataset; bold marks the best value within each backbone--target group.}
\label{tab:ood-gain-appendix}
\begin{tabular}{lllrrrr}
\toprule
\textbf{Backbone} & \textbf{Target} & \textbf{Train source} & \textbf{AUROC} & \textbf{AUPRC} & \textbf{Ret@50} & \textbf{Lift@20} \\
\midrule
Llama3-8B & NQ & ID & 0.779 & 0.851 & 0.575 & 1.501 \\
Llama3-8B & NQ & SQuAD$\rightarrow$NQ & 0.749 & 0.808 & 0.559 & 1.452 \\
Llama3-8B & NQ & TriviaQA$\rightarrow$NQ & 0.756 & 0.822 & 0.563 & 1.447 \\
Llama3-8B & NQ & All$\rightarrow$NQ & \textbf{0.802} & \textbf{0.870} & \textbf{0.595} & \textbf{1.546} \\
\addlinespace[1pt]
Llama3-8B & SQuAD & ID & 0.810 & 0.947 & 0.304 & 1.187 \\
Llama3-8B & SQuAD & NQ$\rightarrow$SQuAD & 0.773 & 0.934 & 0.286 & 1.175 \\
Llama3-8B & SQuAD & TriviaQA$\rightarrow$SQuAD & 0.781 & 0.937 & 0.294 & 1.177 \\
Llama3-8B & SQuAD & All$\rightarrow$SQuAD & \textbf{0.829} & \textbf{0.953} & \textbf{0.311} & \textbf{1.192} \\
\addlinespace[1pt]
Llama3-8B & TriviaQA & ID & 0.922 & 0.862 & 0.960 & 2.977 \\
Llama3-8B & TriviaQA & NQ$\rightarrow$TriviaQA & 0.889 & 0.807 & 0.938 & 2.851 \\
Llama3-8B & TriviaQA & SQuAD$\rightarrow$TriviaQA & 0.895 & 0.798 & 0.946 & 2.819 \\
Llama3-8B & TriviaQA & All$\rightarrow$TriviaQA & \textbf{0.934} & \textbf{0.878} & \textbf{0.975} & \textbf{3.060} \\
\midrule
Qwen3-14B & NQ & ID & 0.810 & 0.912 & 0.473 & 1.354 \\
Qwen3-14B & NQ & SQuAD$\rightarrow$NQ & 0.769 & 0.881 & 0.449 & 1.296 \\
Qwen3-14B & NQ & TriviaQA$\rightarrow$NQ & 0.759 & 0.885 & 0.443 & 1.321 \\
Qwen3-14B & NQ & All$\rightarrow$NQ & \textbf{0.840} & \textbf{0.934} & \textbf{0.498} & \textbf{1.393} \\
\addlinespace[1pt]
Qwen3-14B & SQuAD & ID & 0.755 & 0.896 & 0.396 & 1.266 \\
Qwen3-14B & SQuAD & NQ$\rightarrow$SQuAD & 0.758 & 0.897 & 0.398 & 1.274 \\
Qwen3-14B & SQuAD & TriviaQA$\rightarrow$SQuAD & 0.708 & 0.873 & 0.372 & 1.237 \\
Qwen3-14B & SQuAD & All$\rightarrow$SQuAD & \textbf{0.790} & \textbf{0.916} & \textbf{0.415} & \textbf{1.306} \\
\addlinespace[1pt]
Qwen3-14B & TriviaQA & ID & \textbf{0.894} & \textbf{0.891} & \textbf{0.883} & \textbf{2.292} \\
Qwen3-14B & TriviaQA & NQ$\rightarrow$TriviaQA & 0.821 & 0.801 & 0.808 & 2.097 \\
Qwen3-14B & TriviaQA & SQuAD$\rightarrow$TriviaQA & 0.816 & 0.775 & 0.809 & 2.009 \\
Qwen3-14B & TriviaQA & All$\rightarrow$TriviaQA & 0.893 & 0.887 & 0.872 & 2.285 \\
\bottomrule
\end{tabular}

\end{table*}

\section{Retrieval-Augmented NQ}
\label{sec:appendix_rag}

We include retrieval-augmented NQ as a supplementary setting. In the main
evaluation, NQ serves as a no-retrieval baseline: the model receives the question
but no externally supplied passage. For the retrieval-augmented variants,
contexts are retrieved from SQuAD and TriviaQA corpora rather than provided by
the original benchmark. The BM25 setting uses sparse lexical retrieval over the
candidate contexts, while the neural setting retrieves by dense semantic
similarity and can therefore recover paraphrastic matches that do not share
surface words. We report this setting in the supplementary material because cached baseline
coverage is less complete than in the main CR=0 and CR=1 evaluations.

\begin{table}[t]
\centering
\small
\setlength{\tabcolsep}{4pt}
\caption{Retrieval-augmented NQ statistics.}
\label{tab:rag-dataset-stats}
\resizebox{\columnwidth}{!}{%
\begin{tabular}{lcccc}
\toprule
\textbf{Retrieval} & \textbf{\#Samples} & \textbf{Avg. Len.} &
\textbf{Exact (\%)} & \textbf{Fuzzy (\%)} \\
\midrule
BM25 & 3,610 & 667 & 2.3 & 14.9 \\
Neural & 3,610 & 671 & 5.3 & 21.1 \\
\bottomrule
\end{tabular}
}
\end{table}

\begin{table}[t]
\centering
\footnotesize
\setlength{\tabcolsep}{2.8pt}
\renewcommand{\arraystretch}{1.04}
\caption{Retrieval-augmented NQ results. Results are grouped by retrieval
setting and backbone; bold marks the best value within each group.}
\label{tab:rag-unified-appendix}
\resizebox{\columnwidth}{!}{%
\begin{tabular}{lllrr}
\toprule
\textbf{Set.} & \textbf{Backbone} & \textbf{Method} & \textbf{AUROC} & \textbf{Ret@50} \\
\midrule
NQ & Llama3-8B & HaluNet & \textbf{0.779} & \textbf{0.575} \\
NQ & Llama3-8B & $P(\mathrm{True})$ & 0.692 & 0.524 \\
NQ & Llama3-8B & PE & 0.663 & 0.484 \\
NQ & Llama3-8B & T-NLL & 0.593 & 0.451 \\
\addlinespace[1pt]
NQ & Qwen3-14B & HaluNet & \textbf{0.810} & \textbf{0.473} \\
NQ & Qwen3-14B & $P(\mathrm{True})$ & 0.749 & 0.437 \\
NQ & Qwen3-14B & PE & 0.700 & 0.402 \\
NQ & Qwen3-14B & T-NLL & 0.619 & 0.355 \\
BM25 & Llama3-8B & HaluNet & \textbf{0.770} & \textbf{0.544} \\
BM25 & Llama3-8B & $P(\mathrm{True})$ & 0.611 & 0.428 \\
BM25 & Llama3-8B & PE & 0.643 & 0.439 \\
BM25 & Llama3-8B & T-NLL & 0.555 & 0.402 \\
\addlinespace[1pt]
BM25 & Qwen3-14B & HaluNet & \textbf{0.800} & \textbf{0.436} \\
BM25 & Qwen3-14B & $P(\mathrm{True})$ & 0.668 & 0.346 \\
BM25 & Qwen3-14B & PE & 0.668 & 0.360 \\
BM25 & Qwen3-14B & T-NLL & 0.605 & 0.327 \\
Neural & Llama3-8B & HaluNet & \textbf{0.780} & \textbf{0.524} \\
Neural & Llama3-8B & $P(\mathrm{True})$ & 0.617 & 0.406 \\
Neural & Llama3-8B & PE & 0.632 & 0.420 \\
Neural & Llama3-8B & T-NLL & 0.566 & 0.387 \\
\addlinespace[1pt]
Neural & Qwen3-14B & HaluNet & \textbf{0.831} & \textbf{0.358} \\
Neural & Qwen3-14B & $P(\mathrm{True})$ & 0.681 & 0.283 \\
Neural & Qwen3-14B & PE & 0.694 & 0.289 \\
Neural & Qwen3-14B & T-NLL & 0.606 & 0.258 \\
\bottomrule
\end{tabular}

}
\end{table}
Table~\ref{tab:rag-dataset-stats} shows that neural retrieval has higher exact
and fuzzy answer coverage than BM25, although both retrieval sources remain
noisy. Table~\ref{tab:rag-unified-appendix} shows that HaluNet remains the
strongest method among the reported baselines under both retrieval variants,
indicating that the detector is not limited to the original no-context NQ
setting.

\section{Additional Architecture Analysis}
\label{sec:appendix_arch_scale}

The main-text architecture study focuses on the branch-encoding and fusion
choices that define the final HaluNet model. We omit scale-specific fusion
comparisons from the paper because they do not change the architecture choice:
the Emb-CNN branch encoder with attention fusion gives the strongest
architecture-level result in the full SQuAD CR=0 comparison, while heavier
fusion mechanisms do not provide a consistent enough gain to justify becoming
the default model.

\section{Robustness Checks}
\label{sec:appendix_robustness}

\subsection{Label-Noise Sensitivity}
\label{sec:label_noise}

We evaluate sensitivity to noisy weak supervision by randomly flipping a
fraction of training labels. Table~\ref{tab:label-noise-answer} reports the
available repeated runs on SQuAD with Llama3-8B. We report the AUROC change
relative to the clean-label setting to make the degradation easier to read.
Performance is stable under 5--20\% injected noise and drops more visibly at
30\%, which is consistent with treating judge labels as scalable but imperfect
weak supervision.

\begin{table}[t]
\centering
\footnotesize
\setlength{\tabcolsep}{2.8pt}
\caption{Label-noise sensitivity on SQuAD CR=0 with Llama3-8B. Values after
$\pm$ are standard deviations across repeated seeds; parentheses in the noise
column report AUROC change relative to 0\% noise.}
\label{tab:label-noise-answer}
\resizebox{\columnwidth}{!}{%
\begin{tabular}{lrrrr}
\toprule
\textbf{Noise} & \textbf{Runs} & \textbf{AUROC} & \textbf{Ret@50} & \textbf{Lift@20} \\
\midrule
0\% & 1 & 0.827 & 0.308 & 1.188 \\
5\% (-0.009) & 3 & 0.818$\pm$0.013 & 0.308$\pm$0.004 & 1.189$\pm$0.002 \\
10\% (-0.003) & 3 & 0.824$\pm$0.002 & 0.309$\pm$0.002 & 1.191$\pm$0.002 \\
20\% (-0.008) & 3 & 0.819$\pm$0.001 & 0.306$\pm$0.001 & 1.189$\pm$0.006 \\
30\% (-0.036) & 3 & 0.791$\pm$0.025 & 0.297$\pm$0.010 & 1.179$\pm$0.012 \\
\bottomrule
\end{tabular}

}
\end{table}

\subsection{Maximum Answer Length}
\label{sec:length_sensitivity}

The main experiments use $L=50$ generated tokens. Table~\ref{tab:length-answer}
reports performance under different maximum lengths to check sensitivity to
this truncation choice. Cut denotes the percentage of answers longer than the
maximum length, and Tok. denotes the percentage of generated-answer tokens
retained after truncation. Very short truncation loses useful answer-level
evidence, while longer thresholds give only modest additional gains. The
default $L=50$ covers all answers in this evaluation split.

\begin{table}[t]
\centering
\scriptsize
\setlength{\tabcolsep}{2.2pt}
\caption{Maximum answer length sensitivity on SQuAD CR=0 with Llama3-8B.}
\label{tab:length-answer}
\begin{tabular}{rrrrrrr}
\toprule
\textbf{$L$} & \textbf{Cut (\%)} & \textbf{Tok. (\%)} & \textbf{AUROC} & \textbf{AUPRC} & \textbf{Ret@50} & \textbf{Lift@20} \\
\midrule
3 & 60.0 & 52.5 & 0.804 & 0.943 & 0.298 & 1.180 \\
5 & 28.6 & 71.2 & 0.819 & 0.949 & 0.305 & 1.187 \\
10 & 7.2 & 86.9 & 0.824 & 0.952 & 0.307 & 1.189 \\
25 & 1.5 & 96.1 & 0.825 & 0.951 & 0.306 & 1.184 \\
50 & 0.0 & 100.0 & 0.827 & 0.952 & 0.308 & 1.188 \\
\bottomrule
\end{tabular}

\end{table}

\section{Prompt Templates}
\label{sec:prompts}

\paragraph{Answer generation.}
For answer generation, we use dataset-specific few-shot QA demonstrations when
available and decode with low temperature.

\begin{promptbox}
\begin{lstlisting}
{Context: <context>}
Question: <question>
Answer:
\end{lstlisting}
\end{promptbox}

\paragraph{LLM-as-a-Judge labeling.}
The judge receives the question, reference answer, generated answer, and
context when available, and returns a binary support decision. The decision is
therefore conditioned on explicit evidence from the context or reference answer,
and, when such evidence is unavailable or incomplete, on the judging model's
parametric factual knowledge. We use a compact yes/no template to reduce
formatting variation.

\begin{promptbox}
\begin{lstlisting}
Question: <question>
{Context: <context>}
Reference answer: <reference_answer>
Generated answer: <model_answer>

Is the generated answer supported by the reference answer and available context?
Answer only Yes or No.
\end{lstlisting}
\end{promptbox}

\paragraph{P(True).}
For $P(\mathrm{True})$, the model is prompted to assess whether the generated
answer is supported by the question, reference answer, and available context.
The probabilities of the yes/no responses are converted into hallucination risk
as
\[
S_{P(\mathrm{True})}
=
\frac{p(\mathrm{No})}
{p(\mathrm{Yes})+p(\mathrm{No})+\epsilon}.
\]

\paragraph{SelfCheckGPT.}
For SelfCheckGPT, sampled answers are compared against the candidate answer
using the NLI-based consistency variant in the human-annotated comparison. The
returned inconsistency score is used as hallucination risk, so larger values
indicate higher risk.

\section{Human Annotation Protocol}
\label{sec:human_annotation_protocol}

The human evaluation in Sec.~\ref{sec:rq2_human_ranking} was conducted by
three trained annotators with prior coursework or research experience in natural
language processing. Before annotation, the annotators received written
guidelines and a small calibration set to align their interpretation of
answer-level support. The guidelines instructed annotators to judge whether the
generated answer was supported by the provided question, reference answer, and
available context when present. An answer was marked unsupported if it
contradicted the evidence, introduced an unverifiable entity, number, date, or
condition, answered a different question, or contained a claim that could not be
validated from the supplied materials. Minor wording differences, paraphrases,
and equivalent aliases were treated as supported when they preserved the same
answer semantics.

Each example was annotated independently by all three annotators. The final
binary label was obtained by majority vote. Annotators could additionally mark
an example as ambiguous when the evidence was insufficient, the reference answer
was itself unclear, or the generated answer was only partially supported. These
ambiguous examples were excluded from the human-labeled evaluation set before
computing the reported metrics; this filtering removed 16 examples, leaving 284
valid examples.

The annotation task used benchmark questions, reference answers, contexts, and
model-generated outputs, and did not require annotators to provide personal
information or evaluate sensitive user data. Annotators were informed that their
labels would be used for academic research and agreed to participate before the
task began. They were compensated for their time. No annotator identities are
reported in the paper.

\end{document}